\documentclass[journal]{IEEEtran}
\usepackage{amsmath}
\usepackage{amsfonts}
\usepackage{amssymb}
\usepackage{graphicx}
\usepackage{cite}
\usepackage{float}
\usepackage{multirow}
\usepackage{array}
\usepackage{booktabs}
\usepackage{subfigure}

\usepackage{algorithm}
\usepackage{algorithmicx}
\usepackage{algpseudocode}

\makeatletter
\renewcommand{\maketag@@@}[1]{\hbox{\m@th\normalsize\normalfont#1}}%
\makeatother

\begin{document}
\newenvironment{sequation}{\begin{equation}\scriptsize}{\end{equation}}
\title{Learning from Multiple Time Series: A Deep Disentangled Approach to Diversified Time Series Forecasting}

\author{Ling Chen, Weiqi Chen, Binqing Wu, Youdong Zhang, Bo Wen, Chenghu Yang

\thanks{This work was supported by the National Key Research and Development Program of China under Grant 2018YFB0505000. (Corresponding author: Ling Chen.)}

\thanks{Ling Chen, Weiqi Chen, and Binqing Wu are with the College of Computer Science and Technology, Zhejiang University, Hangzhou 310027, China (emails: lingchen@cs.zju.edu.cn, vc12301@gmail.com, binqingwu@cs.zju.edu.cn).}

\thanks{Youdong Zhang, Bo Wen, and Chenghu Yang are with Alibaba Group, Hangzhou 311100, China (e-mail: \{linqing.zyd, wenbo.wb\}@alibaba-inc.com, yexiang.ych@taobao.com)}}

\markboth{$>$ REPLACE THIS LINE WITH YOUR PAPER IDENTIFICATION NUMBER (DOUBLE-CLICK HERE TO EDIT) $<$}%
{Shell \MakeLowercase{\textit{et al.}}: Bare Demo of IEEEtran.cls for IEEE Journals}
\maketitle

\begin{abstract}
Time series forecasting is a significant problem in many applications, e.g., financial predictions and business optimization. Modern datasets can have multiple correlated time series, which are often generated with global (shared) regularities and local (specific) dynamics. In this paper, we seek to tackle such forecasting problems with DeepDGL, a deep forecasting model that disentangles dynamics into global and local temporal patterns. DeepDGL employs an encoder-decoder architecture, consisting of two encoders to learn global and local temporal patterns, respectively, and a decoder to make multi-step forecasting. Specifically, to model complicated global patterns, the vector quantization (VQ) module is introduced, allowing the global feature encoder to learn a shared codebook among all time series. To model diversified and heterogenous local patterns, an adaptive parameter generation module enhanced by the contrastive multi-horizon coding (CMC) is proposed to generate the parameters of the local feature encoder for each individual time series, which maximizes the mutual information between the series-specific context variable and the long/short-term representations of the corresponding time series. Our experiments on several real-world datasets show that DeepDGL outperforms existing state-of-the-art models.
\end{abstract}

\begin{IEEEkeywords}
Contrastive learning, deep learning, time series forecasting, vector quantization
\end{IEEEkeywords}

\IEEEpeerreviewmaketitle

\section{Introduction}
\IEEEPARstart{T}{ime} series forecasting plays a crucial role in many real-world scenarios, e.g., traffic forecasting \cite{r9}, financial prediction \cite{r7}, and business optimization \cite{r16}. It helps people to make important decisions if the future data values can be predicted accurately. For instance, predicting the future status of urban traffic systems is important for traffic scheduling and management. In a supermarket like Walmart, forecasting the product demand based on historical data can help people do inventory planning to maximize the profit.

\begin{figure}[htbp]
\begin{center}
\includegraphics[width=0.45\textwidth]{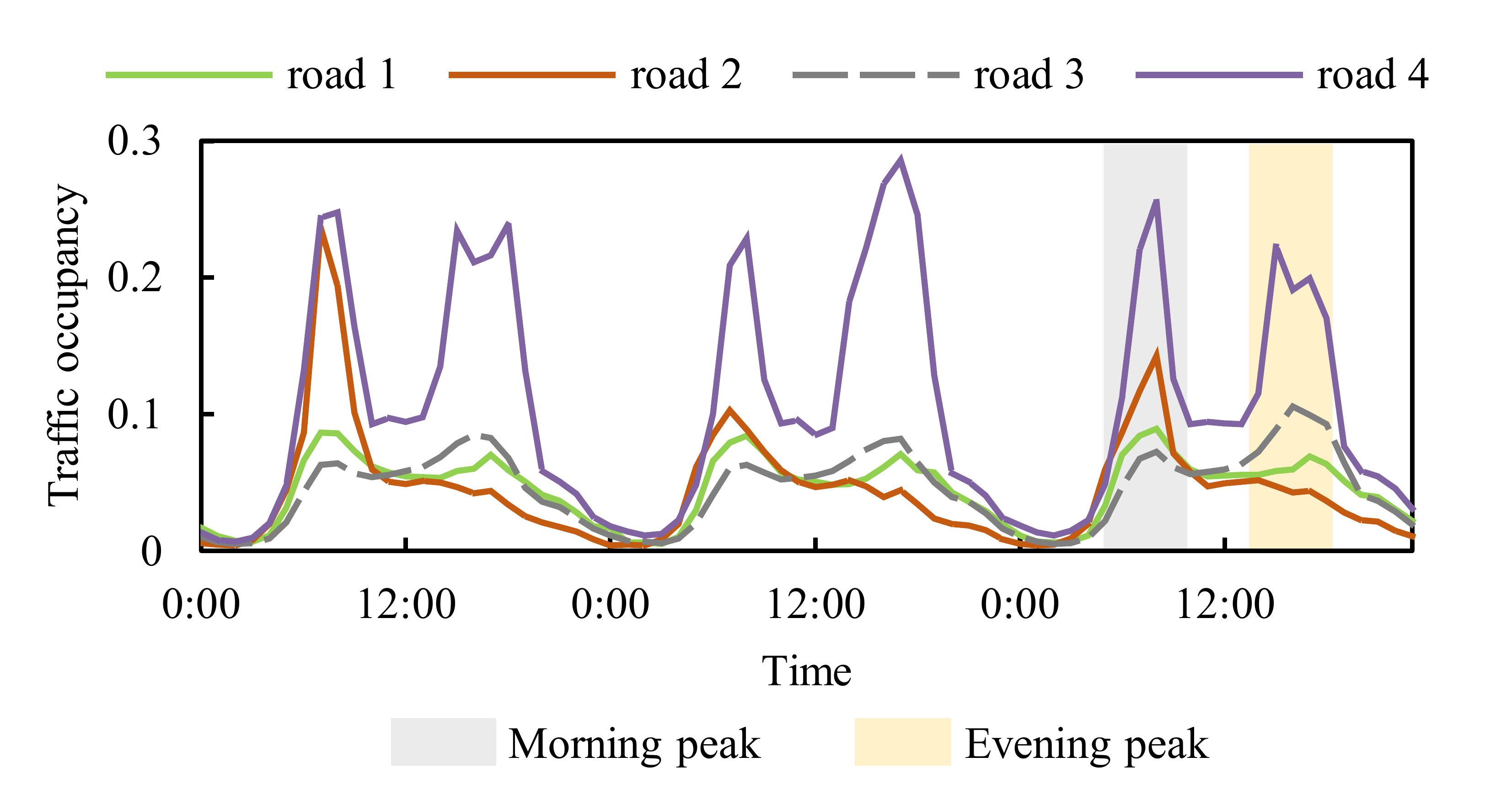}
\end{center}
\caption{Examples of Traffic Occupancy Time Series with Global and Local Temporal Patterns. Global patterns: All time series appear the same period (24 hours) with morning/evening peaks. Local patterns: Road 1 has moderate morning and evening peaks. Road 2 has obvious morning peaks and no evening peaks. Road 3 has slight morning peaks and obvious evening peaks. Road 4 has very intense morning and evening peaks.}
\label{figure_1}
\end{figure}

Real-world time series datasets can have multiple correlated time series, which are often generated with global and local temporal patterns, where global temporal patterns refer to shared regularities among multiple time series, and local temporal patterns indicate dynamics specific to each individual time series (as shown in Figure \ref{figure_1}). In addition, new time series that we are also interested in can be generated, e.g., new products expanded in a supermarket. This leads to a problem of forecasting both the existing and new time series. The main challenges are twofold. First, capturing complicated global temporal patterns as well as diversified and heterogeneous local temporal patterns should be considered jointly. Second, allowing the model to generalize well on new time series is nontrivial, as the dynamics of these unseen data is often different from the existing time series.

Traditional time series forecasting methods are fitted on an individual time series at a time, e.g., AR, ARIMA, exponential smoothing \cite{r10}, and more generally linear state space models (SSMs) \cite{r6}. However, these methods cannot infer shared patterns of multiple time series and benefit from the global information in a dataset. In addition, separately training on each time series makes forecasting time series with short history challenging.

Deep learning based models shed new lights on this problem. Long Short-Term Memory (LSTM) \cite{r5}, Gated Recurrent Units (GRU) \cite{r3}, Temporal Convolution Networks (TCN) \cite{r2}, and Transformer \cite{r8,r18,r23}, have achieved promising results in temporal modeling. These deep models are trained on the entire dataset, utilizing information of all time series, and obtain a shared set of model parameters, which implies the global information. However, these methods cannot explicitly introduce the global information or the information of other related time series when operating on a time series, as they take only the past data of that time series as the input. Moreover, learning shared model parameters has less expressive power to capture the diversified and heterogenous local patterns.

Another route of works tries to incorporate Graph Neural Networks (GNNs) with recurrent neural networks \cite{r1,r9,r24,r25,r26,r27} (e.g., LSTM and GRU) or temporal convolution \cite{r19, r20, r21}, which captures the temporal dependencies and the correlations between different time series jointly. Most of them require a prior graph as inter-series correlations, which are not applicable to the problem faced in this paper. Some methods \cite{r19,r20} introduce an adaptive graph learnt in a data-driven manner to infer the dependencies among different time series. By utilizing GNN, the information of related time series can be introduced when forecasting a time series. However, GNN based methods are helpless to forecast the new time series and tackle datasets with a very large number of time series, as learning such a huge graph is impractical. Furthermore, the shared parameter among all nodes makes them inferior in capturing local temporal patterns accurately.

Recently, some methods, including TRMF \cite{r22} and DeepGLO \cite{r15}, seek to capture global temporal patterns by combining time series forecasting methods with matrix factorization. Each original time series can be represented as a linear combination of $k$ latent time series ($k$ is far less than the number of original time series). However, matrix factorization exploits the global information in the feature space, which is ineffective to capture complicated global temporal patterns. Moreover, due to the shared parameters among all time series, they lack the capacity of modeling highly heterogeneous local temporal patterns.

To this end, we propose DeepDGL, a \underline{deep} forecasting model that \emph{\underline{d}isentangles dynamics into \underline{g}lobal and \underline{l}ocal temporal patterns}. Specifically, DeepDGL follows the encoder-decoder architecture, which has two feature encoders to capture global and local temporal patterns, respectively: 1) in the global feature encoder, the vector quantization (VQ) \cite{r11} module is introduced to maintain a codebook representing global temporal patterns shared by all time series, and an input time series is mapped to the closest global patterns, through which the global information can be introduced; 2) for the local feature encoder, an adaptive parameter generation module enhanced by the \textit{contrastive multi-horizon coding} (CMC) is proposed to generate the parameters of the encoder for each individual time series, which maximizes the mutual information between the series-specific context variable and the long/short-term representations of that time series. Finally, a decoder combines the representations of the global and local feature encoders and performs multi-step forecasting.

During training, existing models (of shared parameters) are only learnt to fit the training data, restricting them to generalize well on new time series, while the adaptive parameter generation module in DeepDGL is learnt to generate specific parameters for each individual time series to model its local temporal patterns. When forecasting new time series, the well-trained parameter generation module can produce suitable parameters for them, endowing DeepDGL with good generalization capacity.

To summarize, our main contributions are as follows:

\begin{itemize}
    \item DeepDGL is proposed, which disentangles dynamics into global temporal patterns shared by multiple time series and local temporal patterns specific to each individual time series.
    \item The VQ module is introduced in the global feature encoder, which learns a codebook representing global temporal patterns to introduce the global information for an input time series.
    \item An adaptive parameter generation module enhanced by CMC is proposed for the local feature encoder, which can effectively capture local information and further generate specific parameters for each time series to model highly diversified and heterogeneous local temporal patterns.
    \item DeepDGL achieves the state-of-the-art performance on three real-world time series datasets and has more powerful generalization capacity. On average, it outperforms the best baseline by 11.8\% on MAPE.
\end{itemize}

\section{Related Works}
Time series forecasting is an emerging topic in machine learning. Traditional methods, e.g., AR, ARIMA, and linear SSMs, mostly operate on an individual time series at a time, which are less helpful for modern datasets with multiple correlated time series. Here, we mainly focus on recent deep learning approaches.

In recent years, deep learning models (cooperated with other methods) are widely applied to time series forecasting, which can be divided into three categories: Classic deep learning methods \cite{r2, r4, r8, r13, r14, r23}, GNN based methods \cite{r1,r9,r19,r20,r21}, and matrix factorization based methods \cite{r15,r22}.

\textbf{Classic deep learning methods} learn a shared set of parameters among all time series in a dataset, while model each individual time series separately. FC-LSTM \cite{r13} forecasts time series with LSTM and fully-connected layers. TCN \cite{r2} introduces causal and dilated convolutions to model temporal patterns. Transformer based methods \cite{r8,r23} take advantage of the self-attention to capture long-term temporal dependencies in the entire time series. DeepAR \cite{r4} is a LSTM based model where the parameters of the distributions of the future values are predicted by LSTM. DeepState \cite{r14} learns the parameters of linear SSMs with deep recurrent neural networks, combining the benefits of the two models.

\textbf{GNN based methods} introduce graph neural networks to capture the correlations between different time series explicitly. Some of them learn adaptive graphs in a data-driven manner without prior graphs as inter-series correlations. Graph WaveNet \cite{r19} combines graph convolutions with an adaptive adjacency matrix and dilated casual convolutions to capture spatial-temporal dependencies. MTGNN \cite{r20} introduces a multi-size convolution module to capture different aspects of temporal dependencies and simplifies the learnt graph to alleviate the overfitting issue. Although most GNN based methods are ineffective to capture local temporal patterns, AGCRN \cite{r1} utilizes a node adaptive parameter learning module to capture node-specific features, which generates node-specific parameters according to the node embeddings.

\textbf{Matrix factorization based methods} represent the original time series as a linear combination of $k$ latent time series capturing global temporal patterns, which are obtained by performing matrix factorization on all time series in a dataset. TRMF \cite{r22} introduces an AR based temporal regularization when factorization. DeepGLO \cite{r15} leverages both global and local features during training and forecasting. The global component introduces a TCN based temporal regularization, and the local component is specialized to focus on the input time series.

Existing methods are ineffective to capture complicated global temporal patterns and heterogenous local temporal patterns and have less generalization capacity on new time series. In this paper, DeepDGL is proposed to address these problems.

\section{Methodology}
\subsection{Problem Setting}
\begin{figure}[htbp]
\centering
\setlength{\abovecaptionskip}{0.cm}
\includegraphics[width=0.37\textwidth]{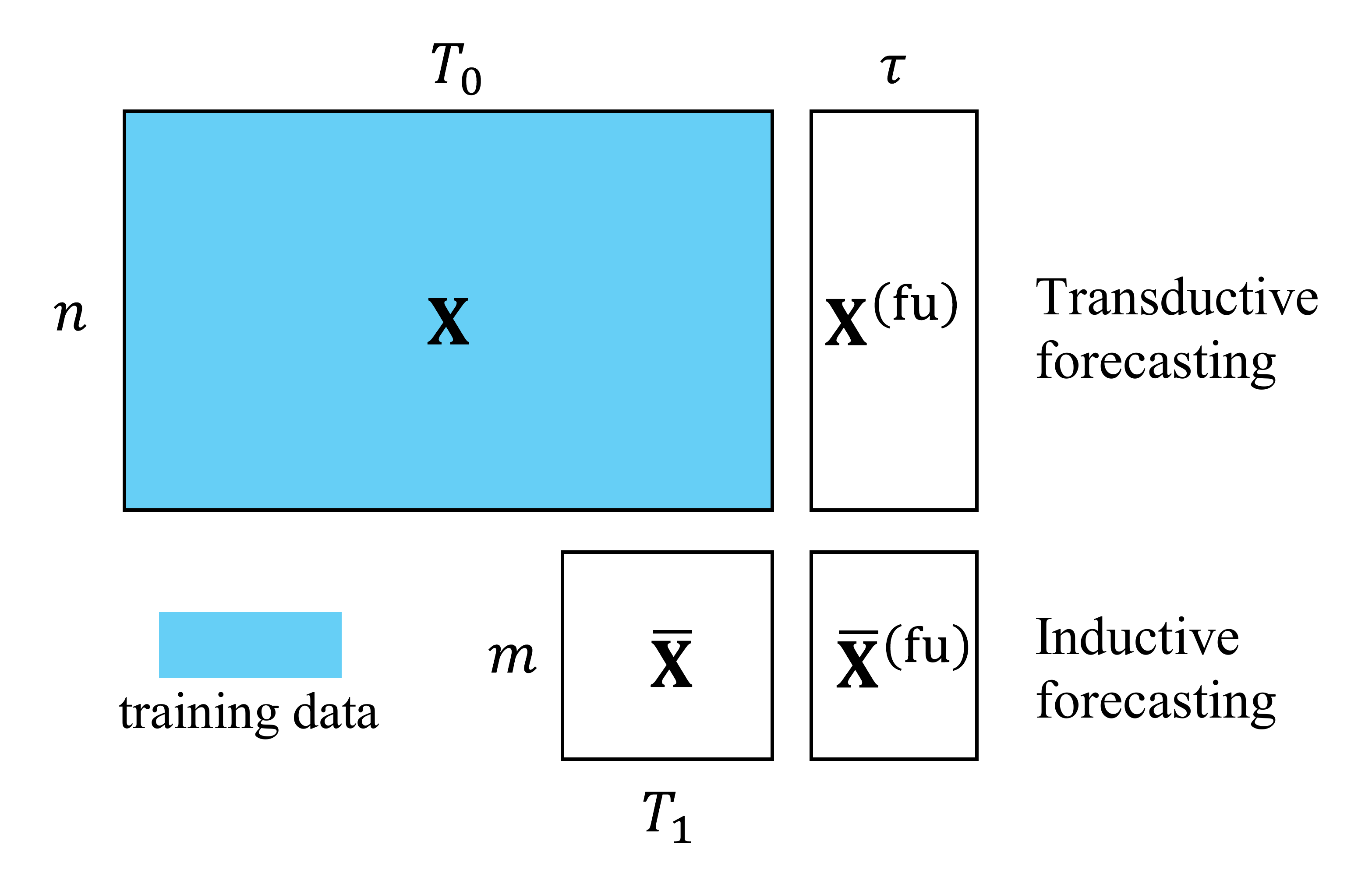}
\caption{Transductive and Inductive Forecasting.}
\label{figure_2}
\end{figure}

Suppose that we have a collection of $n$ correlated time series $\mathbf{X}=\left\{\boldsymbol{x}_{i, 1: T_{0}}\right\}_{i=1}^{n}$ generated from a specific scenario, where $\boldsymbol{x}_{i, 1: T_{0}} \triangleq \left\{x_{i, 1}, x_{i, 2}, \cdots, x_{i, t_{0}}, \cdots, x_{i, T_{0}}\right\}$, $T_0$ is the number of time steps observed, and $x_{i,t_0} \in \mathbb{R}$ denotes the observation of time series $i$ at time step $t_0$. As shown in Figure \ref{figure_2}, using $\mathbf{X}$ as training data, we define two kinds of forecasting tasks:

\textbf{Transductive forecasting.} This task is to predict $\mathbf{X}^{\mathrm{(fu)}} = \left\{\boldsymbol{x}_{i,T_{0}+1:T_{0}+\tau}\right\}_{i=1}^{n}$, the future values of $\mathbf{X}$, where $\tau$ is the prediction horizon.

\textbf{Inductive forecasting.}\footnote{Inductive forecasting is more challenging and requires good generalization of models.} Intuitively, such a data collection can have $m$ new time series, $\overline{\mathbf{X}}=\left\{\overline{\boldsymbol{x}}_{i, 1: T_{1}}\right\}_{i=1}^{m}$, which are correlated with each other and also $\mathbf{X}$. $T_1$ is the number of time steps observed of the new time series, where $T_1 \ll T_0$. It is impractical to train a model using $\overline{\mathbf{X}}$ with short history. This task is to inductively predict $\overline{\mathbf{X}}^{\mathrm{(fu)}} = \left\{\overline{\boldsymbol{x}}_{i,T_{1}+1:T_{1}+\tau}\right\}_{i=1}^{m}$, without training on $\overline{\mathbf{X}}$.

\subsection{Overview}
As mentioned above, existing methods have at least one of the following limitations that prevent them from accurately forecasting diversified time series.
\begin{itemize}
    \item \textbf{Complicated global temporal patterns.} Traditional univariate forecasting models cannot infer shared patterns from a dataset of correlated time series, as they are fitted on each time series separately. Modern deep learning methods learn shared parameters on an entire dataset to exploit the information of all time series without explicitly introducing global information. Some methods utilize matrix factorization to capture the underlying global information in the feature space. However, they are hard to capture complicated global regularities.
    \item \textbf{Heterogenous local temporal patterns.} Most deep learning methods learn a shared set of parameters for all time series, which cannot effectively capture heterogenous local patterns of each individual time series. Training separately on each time series seems to bypass this problem, while training is intractable due to limited training data, and the global information cannot be introduced.
    \item \textbf{Context difference.} The local dynamics of new time series are often different from those of existing time series, which we refer to as context difference. A model is difficult to retrain for new time series with very short history. To our knowledge, existing methods just simply employ the original trained model(s) to these unseen data, which cannot generalize well due to context difference.
\end{itemize}

To better understanding and modeling diversified time series, we introduce three assumptions: a) time series in a dataset are generated with global regularities and local dynamics, where b) global regularities indicate the shared temporal patterns, which can be represented as shared representations in the model, and c) local dynamics are specific temporal patterns of each individual time series, which could be highly heterogenous and diversified, and cannot be effectively captured by a unique set of model parameters.

\begin{figure*}[htbp]
\begin{center}
\includegraphics{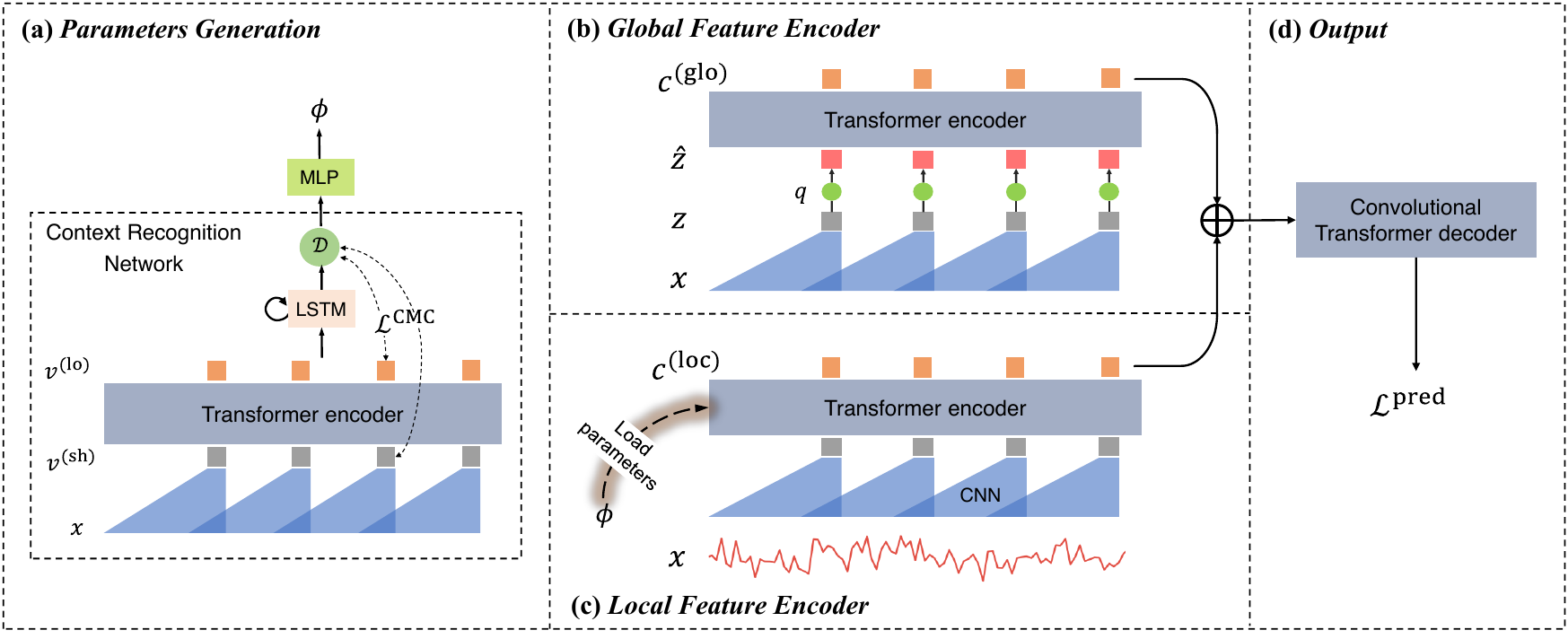}
\end{center}
\caption{Overview of DeepDGL (We only depict 4 time steps of the time series for simplicity). DeepDGL employs an encoder-decoder architecture, with the global feature encoder modeling global temporal patterns, the local feature encoder modeling local temporal patterns, and a convolutional Transformer decoder to make multi-step forecasting. Moreover, an adaptive parameter generation module generates the parameters of the local feature encoder for each individual time series.}
\label{figure_3}
\end{figure*}

\textbf{Model overview.} In light of the above discussions, we propose DeepDGL, a deep forecasting model that disentangles dynamics into global and local temporal patterns. As shown in Figure \ref{figure_3}, DeepDGL employs an encoder-decoder architecture. First, the input time series $\boldsymbol{x}_{1:T}$ are passed through two encoders, capturing global and local patterns, respectively. In the global feature encoder (Figure \ref{figure_3}(b)), the convolution network outputs short-term representations at each time step, and the vector quantization (VQ) module learns a codebook representing global temporal patterns and maps short-term representations at each time step to the vectors in the codebook, which are then fed to the Transformer encoder to output long-term representations at each time step. The local feature encoder also follows the convolutional Transformer encoder (Figure \ref{figure_3}(c)), and its parameters are generated from the series-specific context variable $\mathcal{D}$, which fully captures the local information and is obtained using the \textit{contrastive multi-horizon coding} (CMC) that maximizes mutual information (MI) between $\mathcal{D}$ and long/short-term views of corresponding time series (Figure \ref{figure_3}(a)). Finally, a convolutional transformer decoder is used for multi-step forecasting (Figure \ref{figure_3}(d)).

\subsection{Learning Global Representations}
The global feature encoder (Figure \ref{figure_3}(b)) is designed to capture global temporal patterns, and consists of a multi-layer convolutional feature extractor, which takes the observed time series $\boldsymbol{x}_{1:T}$ as input and outputs short-term representations $\boldsymbol{z}_{1:T}$ at each time step. They are then fed to the vector quantization (VQ) module $q$ to build $\hat{\boldsymbol{z}}_{1:T}$ representing global patterns. Finally, a Transformer encoder is utilized to capture the long-term dependencies of the entire time series and outputs long-term representations $\boldsymbol{c}_{1:T}^{\mathrm{(glo)}}$ at each time step. We choose to use the convolutional Transformer architecture due to its capacity of capturing both long and short-term temporal dependencies. Next, we will introduce the details of each module \footnote{We omit the time step subscript of the variables in the case of no ambiguity.}:

\textbf{Short-term representations with convolutions.} The short-term feature extractor is composed of several blocks containing a causal convolution followed by layer normalization and a ReLU activation. The input time series is normalized to zero mean and unit variance. The stride of the convolution is set to 1, the padding mechanism is used to make sure that the output step size is consistent with the input, and a small-size convolution kernel is used to capture short-term temporal patterns.

\textbf{Global representations with the vector quantization module.} The VQ module is introduced to map the original short-term representations to the vectors representing global patterns. Note that our goal is to disentangle dynamics into global and local components. As a result, in the global feature encoder, we need not to capturing every detail of the dynamics but the global patterns. The VQ module replaces the original representation $\boldsymbol{z}$ by $\hat{\boldsymbol{z}} = \boldsymbol{e}_i$ from a shared codebook $\boldsymbol{e} \in \mathbb{R}^{F \times d}$ among all time series, which contains $F$ vectors of size $d$, regarded as the global representations.

\begin{figure}[htbp]
\begin{center}
\includegraphics[width=0.18\textwidth]{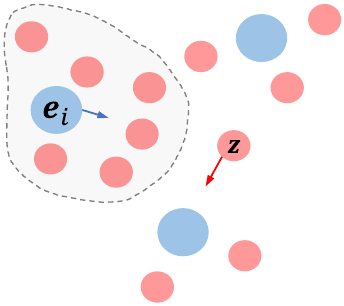}
\end{center}
\caption{The Learning Processes of Global Representations and Short Representations. Blue circles are the global representations in the codebook with the blue arrow as the corresponding learning process. The global representation moves to the short-term representations mapped to it (in the grey dashed ring) and tends to the cluster center of them. Red circles are the short-term representations with the red arrow as the corresponding learning process. The short-term representations are updated to map to appropriate global representations.}
\label{figure_4}
\end{figure}

We choose the global representation $\boldsymbol{e}_i$ from the codebook $\boldsymbol{e}$ by finding the closest one to the input representation $\boldsymbol{z}$ in terms of Euclidean distance, yielding $i=\arg \min _{j}\left\|\boldsymbol{z}-\boldsymbol{e}_{j}\right\|_{2}$. Note that the $\arg \min$ operation is non-differentiable, so we just take the gradients of $\hat{\boldsymbol{z}}$ as those of $\boldsymbol{z}$. During forward-propagation, the nearest representation $\hat{\boldsymbol{z}}$ is passed to the following layers, and during back-propagation, the gradient $\nabla_{\boldsymbol{z}} \mathcal{L}$ is passed unaltered to the preceding convolution networks. Note that due to the straight-through gradient estimation, $\boldsymbol{e}_i$ (i.e., $\hat{\boldsymbol{z}}$) receives no gradient, and we introduce an external objective to learn the global representations in the codebook:
\begin{equation}
\label{r_glo}
\scriptsize
\mathcal{R}^{\mathrm{glo}}=\|\operatorname{sg}(\boldsymbol{z})-\hat{\boldsymbol{z}}\|_{2}^{2}+\gamma\|\boldsymbol{z}-\operatorname{sg}(\hat{\boldsymbol{z}})\|_{2}^{2}
\end{equation}
\noindent where $\operatorname{sg}(x) \equiv x$, $\frac{d}{d x} \operatorname{sg}(x) \equiv 0$ is the stop gradient operator and $\gamma$ is a hyper-parameter. The learning process is visualized in Figure \ref{figure_4}. The first term $\|\operatorname{sg}(\boldsymbol{z})-\hat{\boldsymbol{z}}\|_{2}^{2}$ moves the global representations in the codebook closer to the original short-term representations. As a result, when several short-term representations are mapped to one global representation, the global representation tends to the cluster center of these short-term representations, which makes it more representative. The second term $\|\boldsymbol{z}-\operatorname{sg}(\hat{\boldsymbol{z}})\|_{2}^{2}$ along with the prediction loss $\mathcal{L}^\mathrm{pred}$ urges original short-term representations $\boldsymbol{z}$ to map to appropriate global representations $\hat{\boldsymbol{z}}$ and restricts $\boldsymbol{z}$ to $\hat{\boldsymbol{z}}$.

\textbf{Long-term representations with Transformer.} The output of the VQ module is further fed to a Transformer encoder by taking advantages of the multi-head self-attention mechanism, as self-attention is able to capture long-term dependencies in the entire time series, and multiple attention heads learn to focus on different aspects of temporal patterns. Instead of sine/cosine positional embeddings used in \cite{r18}, a set of time-based covariates (if available) is concatenated to the input representations, e.g., hour-of-the-day, day-of-the-week, which plays a similar role to the positional embeddings. The Transformer encoder is composed of multiple blocks containing a multi-head self-attention layer and a position-wise feedforward network composed of two fully-connected layers (with a ReLU activation in the middle).

\subsection{Generating Local Parameters}
As shown in Figure \ref{figure_3}(c), we also use the convolutional Transformer as the architecture of the local feature encoder, which is identical to the global feature encoder, except the VQ module. As previously mentioned, we need to generate specific parameters of the local feature encoder for each individual time series, as the local dynamics are too heterogenous to be modeled using shared parameters.

To this end, we propose an adaptive parameter generation module (Figure \ref{figure_3}(a)). The first stage is to use the context recognition network (composed of a convolutional Transformer encoder and a LSTM aggregator) to map the input time series to the series-specific context variable $\mathcal{D}$. To make sure that $\mathcal{D}$ fully captures only the local (specific) features of the input time series and filters out the global information, we introduce the \textit{contrastive multi-horizon coding} (CMC) with a KL-divergence regularizer. Then, we input the context variable $\mathcal{D}$ to a two-layer MLP with a ReLU activation in the middle, which outputs the parameters of the local feature encoder. Finally, these parameters are loaded to the encoder. Next, we describe how the CMC help to learn the context variable $\mathcal{D}$.

CMC maximizes the mutual information between the context variable $\mathcal{D}$ (the output of LSTM aggregator) and the short-term representations $\boldsymbol{v}^{(\mathrm{sh})}$ (the output of the convolution module) as well as the mutual information between $\mathcal{D}$ and the long-term representations $\boldsymbol{v}^{(\mathrm{lo})}$ (the output of the transformer encoder) in the context recognition network. Intuitively, the mutual information measures how much better we can infer the value of $\mathcal{D}$ when we know $\boldsymbol{v}^{(\mathrm{sh})}$ or $\boldsymbol{v}^{(\mathrm{lo})}$ than we do not know it. So, maximize the mutual information requires $\mathcal{D}$ to capture the specific high-level information (i.e., long/short term temporal patterns) of the input time series.

Contrastive learning \cite{r12,r17} is widely used in unsupervised learning. Here, CMC introduces two contrastive learning tasks of short-term and long-term representations, and the corresponding objectives are $\mathcal{L}_\mathrm{contrast}^{(\mathrm{sh})}$ and $\mathcal{L}_\mathrm{contrast}^{(\mathrm{lo})}$, respectively. In addition, we also use regularization to ensure that the global information is filtered out.

\textbf{Contrastive loss.} The contrastive learning task requires discriminating the correct representation from the incorrect ones (i.e., distractors). Taking the short-term contrastive task as an example, given the context variable $\mathcal{D}$ and a set ${V}^{(\mathrm{sh})} = \{\boldsymbol{v}_{t}^{(\mathrm{sh})}, \tilde{\boldsymbol{v}}_{1}^{(\mathrm{sh})}, \tilde{\boldsymbol{v}}_{2}^{(\mathrm{sh})}, \cdots, \tilde{\boldsymbol{v}}_{K}^{(\mathrm{sh})}\}$, which contains the short-term representation $\boldsymbol{v}_{t}^{(\mathrm{sh})}$ at time step $t$ (positive representation) of the input series and $K$ distractors $\tilde{\boldsymbol{v}}_{1:K}^{(\mathrm{sh})}$ (negative representations), the model needs to select the correct short-term representation out of the set $V^{(\mathrm{sh})}$. The distractors are uniformly sampled from the short-term representation of other time series. The corresponding loss function is defined as:
\begin{equation}
\label{l_contrast}
\scriptsize
\mathcal{L}_{\mathrm{contrast}}^{(\mathrm{sh})}=\mathbb{E}_{\boldsymbol{x}} \mathbb{E}_{t \sim u(t)}-\log \frac{\exp(f_1(\mathcal{D},\boldsymbol{v}_t^{(\mathrm{sh})};\boldsymbol{x})/ \tau)}{\sum_{\tilde{\boldsymbol{v}}_{t}^{(\mathrm{sh})} \sim V^{(\mathrm{sh})}} \exp (f_1(\mathcal{D}, \tilde{\boldsymbol{v}}_{t}^{(\mathrm{sh})} ; \boldsymbol{x})/\tau)}
\end{equation}
\noindent where $f_1$ is the discrimination function (a 2-layer MLP with a ReLU activation in the middle), which is trained to achieve a high score for positive pairs $(\mathcal{D},\boldsymbol{v}_t^{(\mathrm{sh})})$ and low for negative pairs $(\mathcal{D}, \tilde{\boldsymbol{v}}_{t}^{(\mathrm{sh})})$, $\tau$ is the temperature of SoftMax function, and $u(t)$ is a uniform distribution over time steps $1,2, \cdots,  T$.

$\mathcal{L}_{\mathrm{contrast}}^{(\mathrm{sh})}$ is the categorical cross-entropy of classifying the positive sample correctly from the set ${V}^{(\mathrm{sh})}$ (the first representation is positive), with the predicted probability of: 

\begin{sequation}\frac{\exp(f_1(\mathcal{D},\boldsymbol{v}_t^{(\mathrm{sh})};\boldsymbol{x})/ \tau)}{\sum_{\tilde{\boldsymbol{v}}_{t}^{(\mathrm{sh})} \sim V^{(\mathrm{sh})}} \exp (f_1(\mathcal{D}, \tilde{\boldsymbol{v}}_{t}^{(\mathrm{sh})} ; \boldsymbol{x})/\tau)}
\end{sequation}

The optimal probability for this loss, $p(pos = 0 | V^{(\mathrm{sh})},\mathcal{D})$, should depict that the correct representation $\boldsymbol{v}_{t}^{(\mathrm{sh})}$ comes from the conditional distribution $p(\cdot|\mathcal{D})$ while the distractors $\tilde{\boldsymbol{v}}_{K}^{(\mathrm{sh})}$ come from $p(\cdot)$. Therefore:

\begin{scriptsize}
\begin{equation}
\begin{aligned}
& p(pos = 0 | V^{(\mathrm{sh})},\mathcal{D})\\
&=\frac{p(\boldsymbol{v}_{t}^{(\mathrm{sh})} \mid \mathcal{D}) \prod_{i=1}^{K} p(\widetilde{\boldsymbol{v}}_{i}^{(\mathrm{sh})})}{p(\boldsymbol{v}_{t}^{(\mathrm{sh})} \mid \mathcal{D}) \prod_{i=1}^{K} p(\widetilde{\boldsymbol{v}}_{i}^{(\mathrm{sh})})+\sum_{i=1}^{K} p(\widetilde{\boldsymbol{v}}_{i}^{(\mathrm{sh})} \mid \mathcal{D}) p(\boldsymbol{v}_{t}^{(\mathrm{sh})}) \prod_{j \neq i} p(\widetilde{\boldsymbol{v}}_{j}^{(\mathrm{sh})})}\\
&=\frac{p(\boldsymbol{v}_{t}^{(\mathrm{sh})} \mid \mathcal{D}) / p(\boldsymbol{v}_{t}^{(\mathrm{sh})})}{p(\boldsymbol{v}_{t}^{(\mathrm{sh})} \mid \mathcal{D}) / p(\boldsymbol{v}_{t}^{(\mathrm{sh})})+\sum_{i=1}^{K} p(\widetilde{\boldsymbol{v}}_{i}^{(\mathrm{sh})} \mid \mathcal{D}) / p(\widetilde{\boldsymbol{v}}_{i}^{(\mathrm{sh})})}\\
&=\frac{p(\boldsymbol{v}_{t}^{(\mathrm{sh})} \mid \mathcal{D}) / p(\boldsymbol{v}_{t}^{(\mathrm{sh})})}{\sum_{\widetilde{\boldsymbol{v}}_{t}^{(\mathrm{sh})}{\sim}V^{(\mathrm{sh})}} p(\widetilde{\boldsymbol{v}}_{t}^{(\mathrm{sh})} \mid \mathcal{D}) / p(\widetilde{\boldsymbol{v}}_{t}^{(\mathrm{sh})})}
\end{aligned}
\end{equation}
\end{scriptsize}

By comparing above equation with the predicted probablity, we can find that the optimal $\exp(f_1(\mathcal{D},\boldsymbol{v}_t^{(\mathrm{sh})};\boldsymbol{x})/ \tau)$ is proportional to the density ratio $p(\boldsymbol{v}_{t}^{(\mathrm{sh})} \mid \mathcal{D}) / p(\boldsymbol{v}_{t}^{(\mathrm{sh})})$. Inserting this back in to Equation \ref{l_contrast} (w.r.t. a fixed $t$) results in:

\begin{scriptsize}
\begin{equation}
\begin{aligned}
&\mathcal{L}_{ \mathrm{contrast}}^{(\mathrm{sh})} \\
&=-\mathbb{E}_{\boldsymbol{x}} \log \left[\frac{p\left(\boldsymbol{v}_{t}^{(\mathrm{sh})} \mid \mathcal{D}\right) / p\left(\boldsymbol{v}_{t}^{(\mathrm{sh})}\right)}{p\left(\boldsymbol{v}_{t}^{(\mathrm{sh})} \mid \mathcal{D}\right) / p\left(\boldsymbol{v}_{t}^{(\mathrm{sh})}\right)+\sum_{i=1}^{K} p\left(\widetilde{\boldsymbol{v}}_{i}^{(\mathrm{sh})} \mid \mathcal{D}\right) / p\left(\widetilde{\boldsymbol{v}}_{i}^{(\mathrm{sh})}\right)}\right] \\
&=\mathbb{E}_{\boldsymbol{x}} \log \left[1+\frac{p\left(\boldsymbol{v}_{t}^{(\mathrm{sh})}\right)}{p\left(\boldsymbol{v}_{t}^{(\mathrm{sh})} \mid \mathcal{D}\right)} \sum_{i=1}^{K} \frac{p\left(\widetilde{\boldsymbol{v}}_{i}^{(\mathrm{sh})} \mid \mathcal{D}\right)}{p\left(\widetilde{\boldsymbol{v}}_{i}^{(\mathrm{sh})}\right)}\right] \\
&= \mathbb{E}_{\boldsymbol{x}} \log \left[1+\frac{p\left(\boldsymbol{v}_{t}^{(\mathrm{sh})}\right)}{p\left(\boldsymbol{v}_{t}^{(\mathrm{sh})} \mid \mathcal{D}\right)} K \mathbb{E}_{\tilde{\boldsymbol{v}}_{i}^{(\mathrm{sh})}} \frac{p\left(\widetilde{\boldsymbol{v}}_{i}^{(\mathrm{sh})} \mid \mathcal{D}\right)}{p\left(\widetilde{\boldsymbol{v}}_{i}^{(\mathrm{sh})}\right)}\right] \\
&=\mathbb{E}_{\boldsymbol{x}} \log \left[1+\frac{p\left(\boldsymbol{v}_{t}^{(\mathrm{sh})}\right)}{p\left(\boldsymbol{v}_{t}^{(\mathrm{sh})} \mid \mathcal{D}\right)} K\right] \\
&\geq \mathbb{E}_{\boldsymbol{x}} \log \left[\frac{p\left(\boldsymbol{v}_{t}^{(\mathrm{sh})}\right)}{p\left(\boldsymbol{v}_{t}^{(\mathrm{sh})} \mid \mathcal{D}\right)}(K+1)\right]  \\
&=-M I\left(\mathcal{D} ; \boldsymbol{v}_{t}^{(\mathrm{sh})}\right)+\log (K+1)
\end{aligned}
\end{equation}
\end{scriptsize}

Similarly, given the context variable $\mathcal{D}$ and a set ${V}^{(\mathrm{lo})} = \left\{\boldsymbol{v}_{t}^{(\mathrm{lo})}, \tilde{\boldsymbol{v}}_{1}^{(\mathrm{lo})}, \tilde{\boldsymbol{v}}_{2}^{(\mathrm{lo})}, \cdots, \tilde{\boldsymbol{v}}_{K}^{(\mathrm{lo})}\right\}$, the long-term contrastive loss function is defined as:

\begin{sequation}
\mathcal{L}_{\mathrm{contrast}}^{(\mathrm{lo})}=\mathbb{E}_{\boldsymbol{x}} \mathbb{E}_{t \sim u(t)}-\log \frac{\exp(f_2(\mathcal{D},\boldsymbol{v}_t^{(\mathrm{lo})};\boldsymbol{x})/ \tau)}{\sum_{\tilde{\boldsymbol{v}}_{t}^{(\mathrm{lo})} \sim V^{(\mathrm{lo})}} \exp (f_2(\mathcal{D}, \tilde{\boldsymbol{v}}_{t}^{(\mathrm{lo})} ; \boldsymbol{x})/\tau)}
\end{sequation}

The optimal of the above loss function is: 

\begin{sequation}
\mathcal{L}_{\mathrm{contrast}}^{(\mathrm{lo})} \geq -MI\left(\mathcal{D} ; \boldsymbol{v}_{t}^{(\mathrm{lo})}\right)+\log (K+1)
\end{sequation}

Thus, we have:
\begin{sequation}
\begin{gathered}
MI(\mathcal{D} ; \boldsymbol{x}) \geq \max \left\{{MI}\left(\mathcal{D} ; \boldsymbol{v}_{t}^{(\mathrm{sh})}\right) ; M I\left(\mathcal{D} ; \boldsymbol{v}_{t}^{(\operatorname{lo})}\right)\right\} \geq \\
\frac{1}{2}\left(2 \log (K+1)-\mathcal{L}_{\mathrm{contrast}}^{(\mathrm{lo})}-\mathcal{L}_{\mathrm {contrast}}^{(\mathrm{sh})}\right)
\end{gathered}
\end{sequation}

Hence, minimizing the objectives maximizes a lower bound of $\max \left\{{MI}\left(\mathcal{D} ; \boldsymbol{v}_{t}^{(\mathrm{sh})}\right) ; M I\left(\mathcal{D} ; \boldsymbol{v}_{t}^{(\operatorname{lo})}\right)\right\}$, which is bounded by $MI(\mathcal{D} ; \boldsymbol{x})$ according to the data processing inequality. This urges the context variable $\mathcal{D}$ to capture the local (specific) information of the input time series $\boldsymbol{x}$.

\textbf{Filtering out the global information.} As discussed above, except capturing the local information, $\mathcal{D}$ should also filter out the global information for disentanglement. However, the contrastive tasks cannot help to achieve this goal. In extreme circumstances, when mapping $\boldsymbol{x}$ to $\mathcal{D}$ is a lossless encoding process, $MI(\mathcal{D} ; \boldsymbol{x})$ gets the maximum, and obviously, $\mathcal{D}$ captures both the global and local information of $\boldsymbol{x}$. To this end, we propose to filter out the global information through regularization. On one hand, $\mathcal{D}$ is a low-dimensional vector carrying little information. On the other hand, we introduce a KL regularizer:

\begin{sequation}
\mathcal{L}_{KL}=-KL[q(\mathcal{D} \mid \boldsymbol{x}) \| p(\mathcal{D})]
\end{sequation}

\noindent where $p(\mathcal{D})=\mathcal{N}(0, \mathbf{I})$ and $q(\mathcal{D}|\boldsymbol{x})$ is the posterior distribution of $\mathcal{D}$. This term introduces a prior for $\mathcal{D}$, with the aim of minimizing the amount of information. As a joint result of maximizing $MI(\mathcal{D} ; \boldsymbol{x})$ and minimizing the amount of information in $\mathcal{D}$, the global information will be filtered out, while the local information that can maximize $MI(\mathcal{D} ; \boldsymbol{x})$ is preserved.

Since $\mathcal{D}$ is a stochastic variable, we modify $\mathcal{L}_{\mathrm{contrast}}^{(\mathrm{sh})}$ and $\mathcal{L}_{\mathrm{contrast}}^{(\mathrm{lo})}$, and the final objective of CMC is derived as:

\begin{scriptsize}
\begin{equation}
\label{l_cmc}
\begin{gathered}
\mathcal{L}^{\mathrm{CMC}} =\mathbb{E}_{\boldsymbol{x}}[-\mathbb{E}_{q(\mathcal{D} \mid \boldsymbol{x})}\mathbb{E}_{t \sim u(t)}(\log \frac{\exp (f_{1}(\mathcal{D}, \boldsymbol{v}_{t}^{(\mathrm{sh})} ; \boldsymbol{x}) / \tau)}{\sum_{\tilde{\boldsymbol{v}}_{t}^{(\mathrm{sh})} \sim {V}^{(\mathrm{sh})}} \exp (f_{1}(\mathcal{D}, \boldsymbol{v}_{t}^{(\mathrm{sh})} ; \boldsymbol{x}) / \tau)}\\
+ \log \frac{\exp (f_{2}(\mathcal{D}, \boldsymbol{v}_{t}^{(\mathrm{lo})})/ \tau)}{\sum_{\tilde{\boldsymbol{v}}_{t}^{(\mathrm{lo})} \sim V^{(\mathrm{lo})}}\exp (f_{2}(\mathcal{D}, \boldsymbol{v}_{t}^{(lo)})/\tau)) + \alpha \mathcal{L}_{KL}})]
\end{gathered}
\end{equation}
\end{scriptsize}

\noindent where $\alpha$ is a tunable hyper-parameter.

We found that generating all the parameters of the local feature encoder (convolution networks and the Transformer encoder) is suboptimal, as generating a large number of parameters (the order of magnitude is 1e5) makes the model difficult to train and susceptible to overfitting. We find that sharing network is better. Actually, we generate the parameters (the order of magnitude is 1e4) of the last attention block (a multi-head self-attention layer and a feedforward network), and the remaining parts are shared with the global feature encoder.

\subsection{Learning and Prediction}
\begin{figure}[htbp]
\begin{center}
\includegraphics[width=0.3\textwidth]{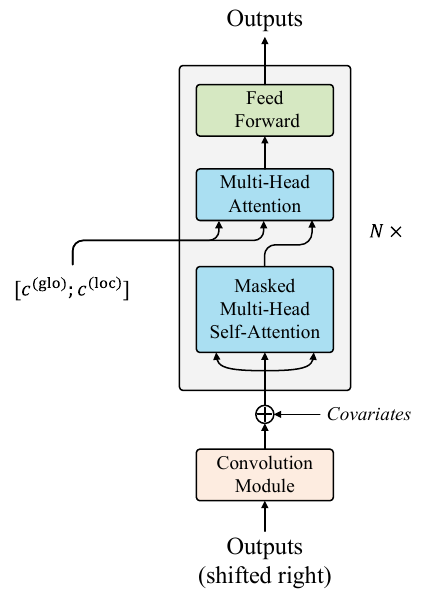}
\end{center}
\caption{The Convolutional Transformer Decoder.}
\label{figure_5}
\end{figure}

Finally, we use the convolutional Transformer decoder for multi-step forecasting (Figure \ref{figure_3}(d)). As shown in Figure \ref{figure_5}, the convolutional Transformer decoder is composed of a convolution module and several identical attention blocks. The convolution module is the same as that in the two encoders, while an attention block includes: (a) a layer of masked multi-head self-attention layer over the outputs of the decoder, where the mask is introduced to prevent time steps from attending to subsequent time steps, (b) a multi-head attention layer over the concatenation of the outputs of the global and local feature encoders, and (c) a position-wise feedforward network composed of two fully-connected layers (with a ReLU activation in the middle).

We use the Mean Absolute Error (MAE) as the prediction loss, which is defined as:

\begin{equation}
\label{l_pred}
\scriptsize
\mathcal{L}^{\mathrm {pred}}=\mathbb{E}_{\boldsymbol{x}} \frac{1}{\tau} \sum_{t=1}^{\tau}\left|\boldsymbol{x}_{T+t}-\boldsymbol{x}_{T+t}^{\mathrm {pred}}\right|
\end{equation}
\noindent where $\boldsymbol{x}_{T+t}$ and $\boldsymbol{x}_{T+t}^{\mathrm {pred}}$ denote the ground truth and prediction at time step $T+t$, respectively, and $\tau$ is the forecasting horizon. The final loss function of DeepDGL can be formulated as a combination of forecasting, CMC, and VQ losses:

\begin{equation}
\label{l}
\scriptsize
\mathcal{L}=\mathcal{L}^{\mathrm{pred}}+\mathcal{L}^{\mathrm{CMC}}+\mathcal{R}^{\mathrm{glo}}
\end{equation}

DeepDGL can be trained in an end-to-end manner using mini-batch gradient descent. When training DeepDGL, we divide the training data $\mathbf{X}$ in the horizontal batch (different time steps) and vertical batch (different time series). When calculating the contrastive multi-horizon coding task, we randomly select the representations of other series in the same mini-batch as negative samples. Algorithm \ref{alg1} gives the detailed training process.

\renewcommand{\algorithmicrequire}{\textbf{Input:}}  
\renewcommand{\algorithmicensure}{\textbf{Output:}} 
\begin{algorithm}[htbp]
\small
  \caption{Mini-Batch Training for DeepDGL}
  \label{alg1} 
  \begin{algorithmic}[1]
    \Require
        training data $\mathbf{X} \in \mathbb{R}^{n \times T_{0}}$, learning rate $\eta$, input horizon $T$, output horizon $\tau$, the number of distractors $K$, horizontal batch size $b_{\mathrm{h}}$, the number of horizontal batches $H$, vertical batch size $b_{\mathrm{v}}$, the number of vertical batches $V$, and the number of epochs $M$.
    \Ensure 
        optimal parameters $\theta$ of DeepDGL
       \State Initialize DeepDGL with parameters $\theta$
       \For{iter $=1,2, \cdots, M$}
            \For{$t=0,b_{\mathrm{h}},2b_{\mathrm{h}},\cdots, (H-1)b_{\mathrm{h}}$}
                \State Select horizontal indices $\mathcal{T}=t,t+1, \cdots, t+b_{\mathrm{h}}-1$
                \For{iter $=1,2, \cdots, V$}
                    \State Randomly select vertical indices $\mathcal{I}=i_{1}, i_{2}, \cdots, i_{b_{\mathrm{v}}}$
                    \State Construct a mini-batch $\mathcal{B}$ of $b_{\mathrm{h}}\times b_{\mathrm{v}}$ samples from $\mathbf{X}$ \Statex \qquad \qquad \ \,using $\mathcal{T}$ and $\mathcal{I}$
                    \For{each sample $\boldsymbol{x}_{1:T},\boldsymbol{x}_{T+1:T+\tau}$ in $\mathcal{B}$ }
                        \State  \textit{/* Forward propagation */}
                        \State Calculate $\boldsymbol{v}_{t}^{(\mathrm{sh})}, \boldsymbol{v}_{t}^{(\mathrm{lo})}, \boldsymbol{z}_{1:T}, \hat{\boldsymbol{z}}_{1:T}$ and $\boldsymbol{x}_{T+1:T+\tau}^{\mathrm{pred}}$
                    \EndFor
                    \State Construct $V^{\mathrm{(sh)}}$ by randomly selecting $\tilde{\boldsymbol{v}}_{i}^{(\mathrm{sh})}{ }_{i=1}^{K}$ in $\mathcal{B}$
                    \State Construct $V^{\mathrm{(lo)}}$ by randomly selecting $\tilde{\boldsymbol{v}}_{i}^{(\mathrm{lo})}{ }_{i=1}^{K}$ in $\mathcal{B}$
                    \State Calculate $\mathcal{L}^{\mathrm{CMC}}$ using Equation \ref{l_cmc}
                    \State Calculate $\mathcal{R}^{\mathrm{glo}}$ using Equation \ref{r_glo}
                    \State Calculate $\mathcal{L}^{\mathrm {pred}}$ using Equation \ref{l_pred}
                    \State Calculate $\mathcal{L}$ using Equation \ref{l}
                    \State  \textit{/* Back propagation */}
                    \State $\theta \leftarrow\left(\theta-\eta \frac{\partial \mathcal{L}}{\partial \theta}\right)$
                \EndFor
            \EndFor
        \EndFor
  \end{algorithmic}
\end{algorithm}


\textbf{Remarks.} In addition to effectively disentangling the dynamics into global and local temporal patterns, DeepDGL also has the following characteristics: (1) During forecasting, although only a single time series is inputted, the codebook in the VQ module implies the information of other time series, which can introduce the global information for it. (2) Instead of training on each individual time series separately to obtain specific parameters (e.g., AR and ARIMA), DeepDGL amortizes learning specific parameters of the local encoder using the adaptive parameter generation module, which exploits correlations between different time series and reduces the learning cost. (3) During training, the inductive bias of the parameter generation module is to learn to generate specific parameters of the local feature encoder for each individual time series to model local temporal patterns. When forecasting new time series, the well-trained parameter generation module can produce suitable parameters for them, endowing DeepDGL with good generalization capacity.

\section{Experiments}
\subsection{Datasets}
We compare the performances of DeepDGL on three real-world datasets, covering energy, traffic, and Internet domains. We use the preprocessed datasets in \cite{r15}. For each dataset, we first randomly select 70\% of the time series, and then employ a sliding window with a step size of 1 on them to construct samples, which are then split in chronological order with 60\% for training, 20\% for transductive validation, and 20\% for transductive testing. For the remaining 10\% and 20\% of the time series, we also employ a sliding window to construct the inductive validation and testing sets \footnote{Samples that overlaps with the training data are not included.}. The statistics of each dataset are shown in Table \ref{table_1}.

\begin{table}
\centering
\caption{Data Statistics}
\label{table_1}
\resizebox{0.48\textwidth}{!}{
\begin{tabular}{llllll} 
\hline
Datasets    & \# of time series & \# of time steps & Granularity\\
\hline
Electricity & 370               & 26136            & 1 hour \\
Traffic     & 963               & 10560            & 1 hour \\
Wiki   & 115,084           & 803              & 1 day\\
\hline
\end{tabular}}
\end{table}

\subsection{Evaluation metrics}
We use the Mean Absolute Percentage Errors (MAPE), Weighted Absolute Percent Error (WAPE), and symmetric Mean Absolute Percentage Errors (SMAPE) to measure the performances, which are averaged by $\tau$ steps in multi-step forecasting \footnote{As the order of magnitude of different time series in a dataset varies, we do not use the Mean Absolute Error (MAE) and Root Mean Squared Error (RMSE) as evaluation metrics}. 

Let $\boldsymbol{X}_t$ and $\hat{\boldsymbol{X}}_t$ denote the ground truth and prediction at time step $t$, respectively. $\tau$ is the forecasting horizon. The evaluation metrics can be computed as follows:
\begin{itemize}
    \item Mean Absolute Percent Error (MAPE):
    \begin{sequation}
    MAPE=\frac{1}{\tau} \sum_{t=1}^{\tau} \frac{|\boldsymbol{X}_{t}-\hat{\boldsymbol{X}}_{t}|}{|\boldsymbol{X}_{t}|}
    \end{sequation}
    \item Weighted Absolute Percent Error (WAPE):
    \begin{sequation}
        WAPE=\frac{\sum_{t=1}^{\tau}|\boldsymbol{X}_{t}-\hat{\boldsymbol{X}}_{t}|}{\sum_{t=1}^{\tau}|\boldsymbol{X}_{t}|}
    \end{sequation}
    \item Symmetric Mean Absolute Percent Error (SMAPE):
    \begin{sequation}
        SMAPE=\frac{1}{\tau} \sum_{t=1}^{\tau} \frac{2|\boldsymbol{X}_{t}-\hat{\boldsymbol{X}}_{t}|}{|\boldsymbol{X}_{t}+\hat{\boldsymbol{X}}_{t}|}
    \end{sequation}
\end{itemize}

\subsection{Settings}
\begin{figure}[htbp]
\begin{center}
\includegraphics[width=0.45\textwidth]{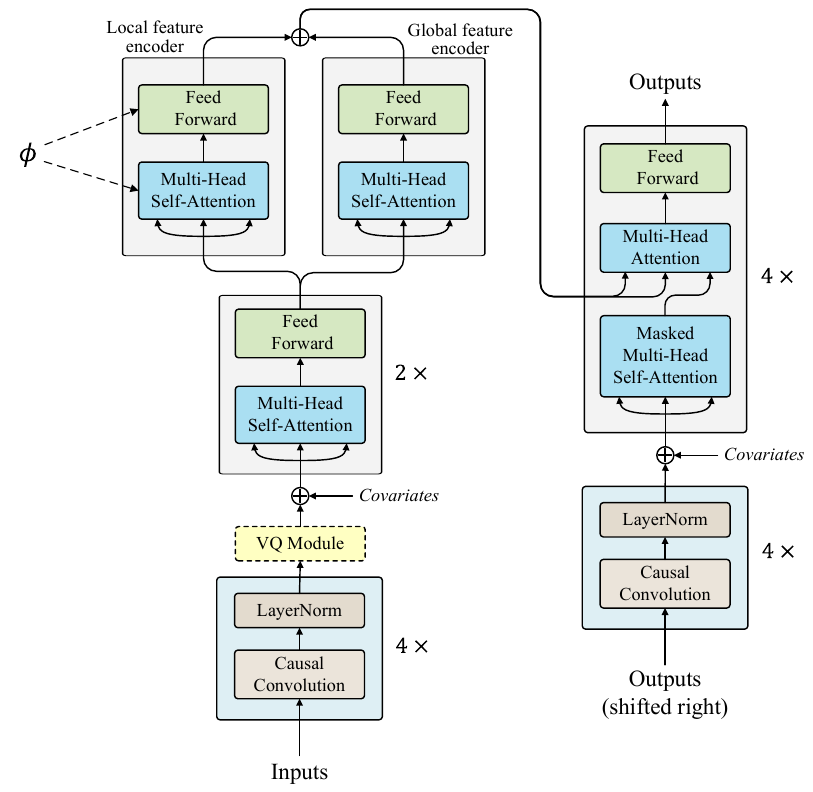}
\end{center}
\caption{The Backbone Network of DeepDGL. The lower-level layers of the global and local feature encoder are shared. The parameters of the last attention block of the local feature encoder are generated by an adaptive parameter generation module. The VQ module is only enabled in the global feature encoder.}
\label{figure_7}
\end{figure}

We use grid search to tune the hyperparameters according to the comprehensive performance of DeepDGL on the validation sets of the two forecasting tasks. The backbone network of DeepDGL is in Figure \ref{figure_7}. On all three of the datasets, we have $\alpha=0.7$ and $\gamma=0.2$, using Adam optimizer for 60 epochs of training. The initial learning rate is 1e-3, and the decay rate is 0.5 for every 10 epochs of training. On the Electricity and Traffic datasets, the horizontal batch size $b_\mathrm{h}$ is set to 32, and the vertical batch size $b_\mathrm{v}$ is set to 64; on the Wiki dataset, $b_\mathrm{h}$ is set to 8, and $b_\mathrm{v}$ is set to 512. Each time series is normalized as $\boldsymbol{x}_{\mathrm{in}}=\left(\boldsymbol{x}_{\mathrm{in}}-\mu\left(\boldsymbol{x}_{\mathrm{in}}\right)\right) / \sigma\left(\boldsymbol{x}_{\mathrm{in}}\right)$, where $\mu$ and $\sigma$ denote the mean and standard deviation, respectively, and the predictions will be scaled back to the original scale, and the metrics are calculated on the original data.

On all datasets, the encoders of DeepDGL includes 4 convolution blocks and 3 attention blocks. The kernel size of convolutions in each layer are: [5, 3, 3, 3], and the feature dimensions are: [64, 64, 64, 64]. The numbers of attention heads are: [4, 4, 4], and the feature dimensions are: [32, 32, 32]. The decoder of DeepDGL contains 4 convolution blocks and 4 attention blocks, where the convolution block is the same as that in the encoders. The numbers of attention heads are: [4, 4, 4, 1], and the feature dimensions are: [32, 32, 32, 1]. 

On the Electricity and Traffic datasets, the number of input steps $T$ and forecasting steps $\tau$ are set to 72 and 24. On the Wiki dataset, $T$ and $\tau$ are set to 42 and 14. On the Electricity and Traffic datasets, $F$, the number of vectors in the codebook, is set to 64; on the Wiki dataset, $F$ is set to 512. With respect to the contrastive multi-horizon coding (CMC), on the Electricity and Traffic datasets, we randomly sample 8 $\boldsymbol{v}_{t}^{(\mathrm{sh})}$ and $\boldsymbol{v}_{t}^{(\mathrm{lo})}$ each (positive samples) for an input time series $\boldsymbol{x}_{1:T}$, and then we randomly sample 32 negative samples from the mini-batch for each positive sample; on the Wiki dataset, constrained by the training time, we randomly sample 4 $\boldsymbol{v}_{t}^{(\mathrm{sh})}$ and $\boldsymbol{v}_{t}^{(\mathrm{lo})}$ each (positive samples) for each input time series $\boldsymbol{x}_{1:T}$, and we randomly sample 8 negative samples from the mini-batch for each positive sample.


\subsection{Baselines}
We compare our model with the following three kinds of baselines:

\textbf{Classic deep learning methods} (FC-LSTM, TCN, DeepAR and DeepState): These methods train a shared model on the entire dataset and employ the original trained model to new data when performing inductive forecasting.
\begin{itemize}
    \item FC-LSTM \cite{r13} forecasts univariate time series using an encoder-decoder architecture. It contains 4 layers for both encoder and decoder with 512 hidden units.
    \item TCN \cite{r2} utilizes dilations and residual connections with the causal convolutions for autoregressive prediction. We stack 5 convolution layers. The dilations are: [1, 2, 4, 8, 16], the feature dimensions are [64, 64, 64, 64, 1], and the kernel size is 5. So, the number of receptive time steps is 125 (which is larger than the number of input time steps of DeepDGL).
    \item DeepAR \cite{r4} is an autoregressive model where the parameters of the distributions for the future values are predicted by LSTM. We use the default hyper-parameters in the GluonTS implementation.
    \item DeepState \cite{r14} combines state space models with deep recurrent neural networks for autoregressive prediction. We use the default hyper-parameters in the GluonTS implementation.
\end{itemize}

\textbf{GNN based methods} (Graph WaveNet and MTGNN): These methods learn adaptive graphs as inter-series dependencies, which cannot scale to large datasets (WiKi) and cannot support inductive forecasting.

\begin{itemize}
    \item Graph WaveNet \cite{r19} introduces an adaptive graph to capture the latent dependencies between time series, and uses dilated convolution to capture temporal dependencies. The configurations and hyper-parameters of this model follow the options recommended in the paper.
    \item MTGNN \cite{r20} uses a graph learning module to learn inter-series correlations and a multi-size dilated convolution module to model temporal dependencies. The configurations and hyper-parameters of this model follow the options recommended in the paper.
\end{itemize}

\textbf{Matrix Factorization based methods} (TRMF and DeepGLO): These methods use matrix factorization to represent each time series as a linear combination of latent time series capturing global patterns. When performing inductive learning, they solve a linear regression to represent input time series with the latent time series learnt on the training data.

\begin{itemize}
    \item TRMF \cite{r22} introduces an autoregressive model to regularize the matrix factorization. The lag indices are set to include the last 3 time steps and the same time step in the last 3 periods.
    \item DeepGLO \cite{r15} has global and local TCNs, which have a same architecture. The dilations are: [1, 2, 4, 8, 16], the feature dimensions are: [32, 32, 32, 32, 1], and the kernel size is 7. So, the number of receptive time steps is 127. 
\end{itemize}

For all autoregressive models, we train them in the whole training data by performing one-step forecasting without explicitly partition samples using a sliding window. During testing, we perform $\tau$-step forecasting and calculate errors, and the number of receptive time steps is consistent with that during training.

\subsection{Comparisons}

\begin{table*}
\centering
\caption{Transductive Forecasting Results on Different Datasets.}
\label{table_2}
\begin{tabular}{lccccccccc} 
\hline
\multirow{2}{*}{Models} & \multicolumn{3}{c}{Electricity}                  & \multicolumn{3}{c}{Traffic}             & \multicolumn{3}{c}{Wiki}                          \\ 
\cline{2-10}
                        & MAPE           & WAPE           & SMAPE          & MAPE           & WAPE           & SMAPE & MAPE           & WAPE           & SMAPE           \\ 
\hline
FC-LSTM                 & 0.354          & 0.214          & 0.238          & 0.278          & 0.127          & 0.189 & 0.595          & 0.644          & 0.517           \\
TCN                     & 0.342          & 0.220          & 0.204          & 0.272          & 0.148          & 0.126 & 0.559          & 0.455          & 0.476           \\
DeepAR                  & 0.296          & \textbf{0.121} & 0.197          & 0.202          & 0.152          & 0.114 & 0.554          & 0.393          & 0.475           \\
DeepState               & 0.337          & 0.215          & 0.265          & 0.264          & 0.133          & 0.119 & 0.612          & 0.485          & 0.395           \\
Graph WaveNet           & 0.283          & 0.255          & 0.211          & 0.223          & 0.114          & 0.127 & /              & /              & /               \\
MTGNN                   & 0.289          & 0.207          & 0.280          & 0.174          & \textbf{0.087} & \textbf{0.098} & /              & /              & /               \\
TRMF                    & 0.329          & 0.231          & 0.246          & 0.242          & 0.132          & 0.160 & 0.630          & 0.533          & 0.510           \\
DeepGLO                 & 0.303          & 0.144          & 0.210          & 0.228          & 0.092          & 0.129 & \textbf{0.474} & 0.485          & 0.368           \\
DeepDGL                 & \textbf{0.261} & 0.152          & \textbf{0.194} & \textbf{0.167} & 0.094          & \textbf{0.098} & 0.528          & \textbf{0.377} & \textbf{0.348}  \\
\hline
\end{tabular}
\end{table*}

\begin{table*}
\centering
\caption{Inductive Forecasting Results on Different Datasets.}
\label{table_3}
\begin{tabular}{lccccccccc} 
\hline
\multirow{2}{*}{Models} & \multicolumn{3}{c}{Electricity}                  & \multicolumn{3}{c}{Traffic}                      & \multicolumn{3}{c}{Wiki}                          \\ 
\cline{2-10}
                        & MAPE           & WAPE           & SMAPE          & MAPE           & WAPE           & SMAPE          & MAPE           & WAPE           & SMAPE           \\ 
\hline
FC-LSTM                 & 0.507          & 0.420          & 0.433          & 0.425          & 0.221          & 0.368          & 0.692          & 0.704          & 0.781           \\
TCN                     & 0.519          & 0.365          & 0.398          & 0.516          & 0.319          & 0.354          & 0.599          & 0.467          & 0.658           \\
DeepAR                  & 0.432          & 0.274          & \textbf{0.322} & 0.470          & 0.289          & 0.226          & 0.746          & 0.567          & 0.520           \\
DeepState               & 0.458          & 0.314          & 0.419          & 0.394          & 0.301          & 0.270          & 0.697          & 0.675          & 0.581           \\
Graph WaveNet           & /              & /              & /              & /              & /              & /              & /              & /              & /               \\
MTGNN                   & /              & /              & /              & /              & /              & /              & /              & /              & /               \\
TRMF                    & 0.488          & 0.455          & 0.356          & 0.451          & 0.269          & 0.356          & 0.577          & 0.558          & 0.543           \\
DeepGLO                 & 0.409          & 0.277          & 0.333          & 0.354          & 0.252          & 0.247          & \textbf{0.451} & 0.461          & 0.360           \\
\textbf{DeepDGL}        & \textbf{0.381} & \textbf{0.247} & 0.353          & \textbf{0.259} & \textbf{0.187} & \textbf{0.166} & 0.525          & \textbf{0.454} & \textbf{0.423}  \\
\hline
\end{tabular}
\end{table*}

Tables \ref{table_2} and \ref{table_3} show the performances of DeepDGL and the baseline models in transductive and inductive forecasting on three datasets, respectively. We can draw the following conclusions:

DeepDGL is superior to the three categories of baselines and lowers down the forecasting error by 9.7\% and 11.4\% over the best baseline in transductive and inductive forecasting, respectively. DeepDGL has significant advantages over classic deep learning methods (LSTM, TCN, DeepAR, and DeepState), as it introduces the VQ module to learn global representations, which can explicitly introduce the global information. Moreover, it has a local feature encoder with specific parameters for each individual time series, which improves the capacity of modeling local temporal patterns.

GNN based methods (Graph WaveNet and MTGNN) outperform classic deep learning methods, which indicates that explicitly learning the correlations between time series using an adaptive graph is useful. By using GNN, the model can introduce the information of related time series when forecasting one time series, which has a similar effect to capturing global information. Unfortunately, GNN based methods cannot scale to large datasets or inductive forecasting. In addition, they also suffer from ineffectively modeling highly heterogenous local patterns, as parameters are shared among all nodes in GNN. Therefore, their performance is inferior to that of DeepDGL.

DeepDGL is also significantly improved compared to DeepGLO, empirically justifying the superiority of DeepDGL, which captures complicated global information in representation space by using nonlinear deep neural networks. Meanwhile, the local feature encoder with specific parameters can more effectively model the heterogeneous local temporal patterns. TRMF does not achieve satisfactory performance. The reason might be that TRMF pays no attention on the input time series, but only relies on the latent time series.

Moreover, DeepDGL has more obvious advantages than all baselines in inductive forecasting, which is a more challenging task and requires good generalization capacity of models, as the new time series tend not to have consistent temporal patterns with the training time series. Classic deep learning methods directly employ the trained model to the unseen data is naive. Matrix Factorization based methods benefit from introducing global information for the new time series through the latent time series. However, due to sharing parameters, they are less effective to model the local temporal patterns of these new time series, which prevents them from generalizing well on new time series. It is convincing that the adaptive parameter generation module in DeepDGL can produce suitable parameters for new time series, endowing DeepDGL with good generalization.

\subsection{Ablation Study}
To verify the effectiveness of each module in DeepDGL, we design four variants: (1) ConvTransformer: Convolutional Transformer encoder-decoder model, which is the basic structure of DeepDGL; (2) w/o CMC: directly generate the parameters of local feature encoder without using CMC; (3) Global-Only: only global feature encoder is used; (4) Local-Only: only local feature encoder is used.

\begin{table*}
\centering
\caption{The Performance of DeepDGL and Its Variants on the Electricity Dataset.}
\label{table_4}
\begin{tabular}{lcccccc} 
\hline
Tasks                                     & Metrics & DeepDGL & ConvTransformer & w/o CMC & Global-Only & Local-Only  \\ 
\hline
\multirow{3}{*}{Transductive Forecasting} & MAPE    & \textbf{0.261}   & 0.304           & 0.283   & 0.319       & 0.294       \\
                                          & WAPE    & \textbf{0.152}   & 0.233           & 0.216   & 0.252       & 0.241       \\
                                          & SMAPE   & \textbf{0.194}   & 0.240           & 0.239   & 0.236       & 0.219       \\ 
\hline
\multirow{3}{*}{Inductive Forecasting}    & MAPE    & \textbf{0.381}   & 0.462           & 0.440   & 0.505       & 0.417       \\
                                          & WAPE    & \textbf{0.247}   & 0.355           & 0.364   & 0.411       & 0.317       \\
                                          & SMAPE   & \textbf{0.353}   & 0.427           & 0.402   & 0.407       & 0.398       \\
\hline
\end{tabular}
\end{table*}

Table \ref{table_4} shows the performance of DeepDGL and its variants on the Electricity dataset for transductive and inductive forecasting. The result indicates that each module in DeepDGL is indispensable. To be Specific:

DeepDGL is superior to Local-Only, indicating that the global representations learnt in the global feature encoder is useful and can effectively capture the global temporal patterns.

DeepDGL outperforms w/o CMC and Global-Only, indicating that learning a specific set of parameters for each individual time series can improve the capacity of modeling local temporal patterns, and introducing CMC can further capture the local information of time series, thus generating more effective model parameters.

In addition, we also found some other phenomena: (1) ConvTranformer is slightly better than LSTM and TCN (Tables 2 and 3), indicating that introducing convolution networks and self-attention mechanism to model short-term and long-term dependencies is effective, and this is why we chose it as the basic network of DeepDGL; (2) Local-Only is significantly better than ConvTranformer in inductive forecasting, while the difference is smaller in transductive forecasting. Note that the only difference between the two variants is that Local-Only uses the adaptive parameter generation module to generate specific parameters of the local feature encoder for each time series, while ConvTranformer learns shared parameters for all time series. This indicates that the adaptive parameter generation module can help the model capture heterogeneous local temporal patterns and thus generalize on new time series better.

\subsection{Case Study}
To investigate the capacity of modeling diversified and heterogenous local temporal patterns, we perform a case study using the Wiki dataset in inductive learning. Time series in this dataset appear highly heterogenous local temporal patterns.

\begin{figure}
\centering
\subfigure[DeepDGL forecasts the downtrend.]{
\includegraphics[width=0.45\textwidth]{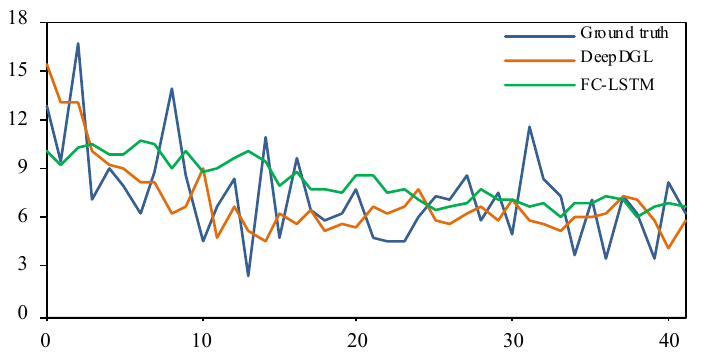}
\label{figure_6a}
}

\subfigure[DeepDGL captures the periodicity.]{
\includegraphics[width=0.45\textwidth]{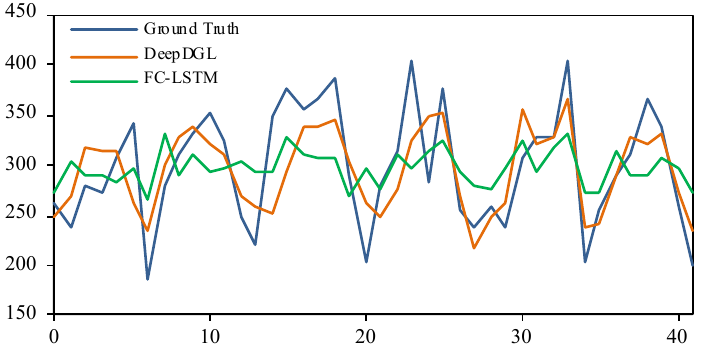}
\label{figure_6b}
}
\caption{Results of Wikipedia Articles Traffic Forecasting.}
\label{figure_6}
\end{figure}

Figure \ref{figure_6} illustrates the 7 days ahead forecasting results of the traffic of two articles. Specifically, we take the predicted values of the 7th day of the multi-step forecasting for visualization. We can find that the predictions of DeepDGL is more consistent with the ground truth. As shown in Figure \ref{figure_6a}, the website traffic decreases, and DeepDGL successfully captures such a local temporal pattern in advance of a week. However, baseline models (e.g., FC-LSTM) fail to capture it. This is because DeepDGL learns a specific set of parameters for each individual time series, which effectively improves the model capacity of capturing local patterns. As shown in Figure \ref{figure_6b}, DeepDGL captures the periodicity more accurately. This also reflects that DeepDGL has effectiveness of tackling diversified time series with heterogenous local patterns.

\section{Conclusions and Future Work}
We propose DeepDGL, which can effectively disentangle the dynamics of time series into global and local temporal patterns. When modeling complicated global temporal patterns, the vector quantization module is introduced to learn a codebook representing global information. When modeling heterogenous local temporal patterns, an adaptive parameter generation module is proposed to generate specific parameters for each individual time series. Our model has achieved the state-of-the-art performance on three real-world datasets.

For future work, we will investigate the following two aspects: (1) How the external information (e.g., POIs in traffic forecasting), can help the model disentangle the dynamics better; (2) Whether generating time-variant parameters of the local feature encoder can help better capture complicated local temporal patterns.

\bibliographystyle{IEEEtran}
\bibliography{references.bib}

\begin{thebibliography}{10}
\providecommand{\url}[1]{#1}
\csname url@samestyle\endcsname
\providecommand{\newblock}{\relax}
\providecommand{\bibinfo}[2]{#2}
\providecommand{\BIBentrySTDinterwordspacing}{\spaceskip=0pt\relax}
\providecommand{\BIBentryALTinterwordstretchfactor}{4}
\providecommand{\BIBentryALTinterwordspacing}{\spaceskip=\fontdimen2\font plus
\BIBentryALTinterwordstretchfactor\fontdimen3\font minus
  \fontdimen4\font\relax}
\providecommand{\BIBforeignlanguage}[2]{{%
\expandafter\ifx\csname l@#1\endcsname\relax
\typeout{** WARNING: IEEEtran.bst: No hyphenation pattern has been}%
\typeout{** loaded for the language `#1'. Using the pattern for}%
\typeout{** the default language instead.}%
\else
\language=\csname l@#1\endcsname
\fi
#2}}
\providecommand{\BIBdecl}{\relax}
\BIBdecl

\bibitem{r9}
Y.~Li, R.~Yu, C.~Shahabi, and Y.~Liu, ``Diffusion convolutional recurrent
  neural network: Data-driven traffic forecasting,'' in \emph{6th International
  Conference on Learning Representations}, 2018.

\bibitem{r7}
B.~Gui, X.~Wei, Q.~Shen, J.~Qi, and L.~Guo, ``Financial time series forecasting
  using support vector machine,'' in \emph{Proceedings of the 10th
  International Conference on Computational Intelligence and Security}, 2014,
  pp. 39--43.

\bibitem{r16}
M.~W. Seeger, D.~Salinas, and V.~Flunkert, ``Bayesian intermittent demand
  forecasting for large inventories,'' in \emph{Proceedings of the 30th
  Conference on Neural Information Processing Systems}, vol.~29, 2016, p.
  4646–4654.

\bibitem{r10}
E.~McKenzie, ``General exponential smoothing and the equivalent {ARMA}
  process,'' \emph{Journal of Forecasting}, vol.~3, no.~3, pp. 333--344, 1984.

\bibitem{r6}
R.~Hyndman, A.~Koehler, J.~Ord, and R.~Snyder, \emph{Forecasting with
  Exponential Smoothing: The State Space Approach}, ser. Springer Series in
  Statistics.\hskip 1em plus 0.5em minus 0.4em\relax Springer Berlin
  Heidelberg, 2008.

\bibitem{r5}
F.~Gers, J.~Schmidhuber, and F.~Cummins, ``Learning to forget: Continual
  prediction with {LSTM},'' in \emph{Proceedings of the 9th International
  Conference on Artificial Neural Networks}, vol.~2, 1999, pp. 850--855.

\bibitem{r3}
\BIBentryALTinterwordspacing
J.~Chung, {\c{C}}.~G{\"{u}}l{\c{c}}ehre, K.~Cho, and Y.~Bengio, ``Empirical
  evaluation of gated recurrent neural networks on sequence modeling,''
  \emph{CoRR}, vol. abs/1412.3555, 2014. [Online]. Available:
  \url{http://arxiv.org/abs/1412.3555}
\BIBentrySTDinterwordspacing

\bibitem{r2}
\BIBentryALTinterwordspacing
S.~Bai, J.~Z. Kolter, and V.~Koltun, ``An empirical evaluation of generic
  convolutional and recurrent networks for sequence modeling,'' \emph{CoRR},
  vol. abs/1803.01271, 2018. [Online]. Available:
  \url{http://arxiv.org/abs/1803.01271}
\BIBentrySTDinterwordspacing

\bibitem{r8}
S.~Li, X.~Jin, Y.~Xuan, X.~Zhou, W.~Chen, Y.-X. Wang, and X.~Yan, ``Enhancing
  the locality and breaking the memory bottleneck of transformer on time series
  forecasting,'' in \emph{Proceedings of the 33rd Conference on Neural
  Information Processing Systems}, vol.~32, 2019.

\bibitem{r18}
A.~Vaswani, N.~Shazeer, N.~Parmar, J.~Uszkoreit, L.~Jones, A.~N. Gomez,
  {\L}.~Kaiser, and I.~Polosukhin, ``Attention is all you need,'' in
  \emph{Proceedings of the 31st Conference on Neural Information Processing
  Systems}, vol.~30, 2017, p. 5998–6008.

\bibitem{r23}
H.~Zhou, S.~Zhang, J.~Peng, S.~Zhang, J.~Li, H.~Xiong, and W.~Zhang,
  ``Informer: Beyond efficient transformer for long sequence time-series
  forecasting,'' in \emph{Proceedings of the 35th AAAI Conference on Artificial
  Intelligence}, vol.~35, 2021, pp. 11\,106--11\,115.

\bibitem{r1}
L.~Bai, L.~Yao, C.~Li, X.~Wang, and C.~Wang, ``Adaptive graph convolutional
  recurrent network for traffic forecasting,'' in \emph{Proceedings of the 34th
  Conference on Neural Information Processing Systems}, vol.~33, 2020, pp.
  17\,804--17\,815.

\bibitem{r24}
W.~Chen, L.~Chen, Y.~Xie, W.~Cao, Y.~Gao, and X.~Feng, ``Multi-range attentive
  bicomponent graph convolutional network for traffic forecasting,'' in
  \emph{Proceedings of the 34th AAAI Conference on Artificial Intelligence},
  vol.~34, 2020, pp. 3529--3536.

\bibitem{r25}
M.~Lv, Z.~Hong, L.~Chen, T.~Chen, T.~Zhu, and S.~Ji, ``Temporal multi-graph
  convolutional network for traffic flow prediction,'' \emph{IEEE Transactions
  on Intelligent Transportation Systems}, vol.~22, no.~6, pp. 3337--3348, 2021.

\bibitem{r26}
\BIBentryALTinterwordspacing
L.~Chen, J.~Xu, B.~Wu, Y.~Qian, Z.~Du, Y.~Li, and Y.~Zhang, ``Group-aware graph
  neural network for nationwide city air quality forecasting,'' \emph{CoRR},
  vol. abs/2108.12238, 2021. [Online]. Available:
  \url{http://arxiv.org/abs/2108.12238}
\BIBentrySTDinterwordspacing

\bibitem{r27}
\BIBentryALTinterwordspacing
J.~Xu, L.~Chen, M.~Lv, C.~Zhan, S.~Chen, and J.~Chang, ``{HighAir}: A
  hierarchical graph neural network-based air quality forecasting method,''
  \emph{CoRR}, vol. abs/2101.04264, 2021. [Online]. Available:
  \url{http://arxiv.org/abs/2101.04264}
\BIBentrySTDinterwordspacing

\bibitem{r19}
Z.~Wu, S.~Pan, G.~Long, J.~Jiang, and C.~Zhang, ``Graph wavenet for deep
  spatial-temporal graph modeling,'' in \emph{Proceedings of the 28th
  International Joint Conference on Artificial Intelligence}, 2019, pp.
  1907--1913.

\bibitem{r20}
Z.~Wu, S.~Pan, G.~Long, J.~Jiang, X.~Chang, and C.~Zhang, ``Connecting the
  dots: Multivariate time series forecasting with graph neural networks,'' in
  \emph{Proceedings of the 26th ACM SIGKDD International Conference on
  Knowledge Discovery \&; Data Mining}, 2020, p. 753–763.

\bibitem{r21}
B.~Yu, H.~Yin, and Z.~Zhu, ``Spatio-temporal graph convolutional networks: A
  deep learning framework for traffic forecasting,'' in \emph{Proceedings of
  the 27th International Joint Conference on Artificial Intelligence}, 2018,
  pp. 3634--3640.

\bibitem{r22}
H.-F. Yu, N.~Rao, and I.~S. Dhillon, ``Temporal regularized matrix
  factorization for high-dimensional time series prediction,'' in
  \emph{Proceedings of the 30th Conference on Neural Information Processing
  Systems}, vol.~29, 2016, p. 847–855.

\bibitem{r15}
R.~Sen, H.-F. Yu, and I.~S. Dhillon, ``Think globally, act locally: A deep
  neural network approach to high-dimensional time series forecasting,'' in
  \emph{Proceedings of the 33rd Conference on Neural Information Processing
  Systems}, vol.~32, 2019, p. 4838–4847.

\bibitem{r11}
A.~van~den Oord, O.~Vinyals, and koray Kavukcuoglu, ``Neural discrete
  representation learning,'' in \emph{Proceedings of the 31st Conference on
  Neural Information Processing Systems}, vol.~30, 2017, p. 6306–6315.

\bibitem{r4}
\BIBentryALTinterwordspacing
V.~Flunkert, D.~Salinas, and J.~Gasthaus, ``{DeepAR}: Probabilistic forecasting
  with autoregressive recurrent networks,'' \emph{CoRR}, vol. abs/1704.04110,
  2017. [Online]. Available: \url{http://arxiv.org/abs/1704.04110}
\BIBentrySTDinterwordspacing

\bibitem{r13}
S.~S. Rangapuram, M.~W. Seeger, J.~Gasthaus, L.~Stella, Y.~Wang, and
  T.~Januschowski, ``Deep state space models for time series forecasting,'' in
  \emph{Proceedings of the 32nd Conference on Neural Information Processing
  Systems}, vol.~31, 2018, p. 7796–7805.

\bibitem{r14}
I.~Sutskever, O.~Vinyals, and Q.~V. Le, ``Sequence to sequence learning with
  neural networks,'' in \emph{Proceedings of the 28th Conference on Neural
  Information Processing Systems}, vol.~27, 2014, p. 3104–3112.

\bibitem{r12}
\BIBentryALTinterwordspacing
A.~van~den Oord, Y.~Li, and O.~Vinyals, ``Representation learning with
  contrastive predictive coding,'' \emph{CoRR}, vol. abs/1807.03748, 2018.
  [Online]. Available: \url{http://arxiv.org/abs/1807.03748}
\BIBentrySTDinterwordspacing

\bibitem{r17}
\BIBentryALTinterwordspacing
Y.~Tian, D.~Krishnan, and P.~Isola, ``Contrastive multiview coding,''
  \emph{CoRR}, vol. abs/1906.05849, 2019. [Online]. Available:
  \url{http://arxiv.org/abs/1906.05849}
\BIBentrySTDinterwordspacing

\end{thebibliography}

\begin{IEEEbiography}[{\includegraphics[width=1in,height=1.25in,clip,keepaspectratio]{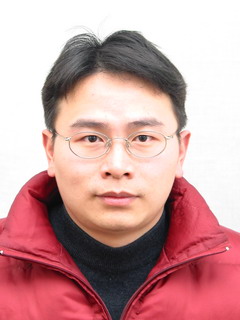}}]
{Ling Chen} received his B.S. and Ph.D. degrees in computer science from Zhejiang University, China, in 1999 and 2004, respectively. He is currently a professor with the College of Computer Science and Technology, Zhejiang University, China. His research interests include ubiquitous computing and data mining.
\end{IEEEbiography}

\begin{IEEEbiography}[{\includegraphics[width=1in,height=1.25in,clip,keepaspectratio]{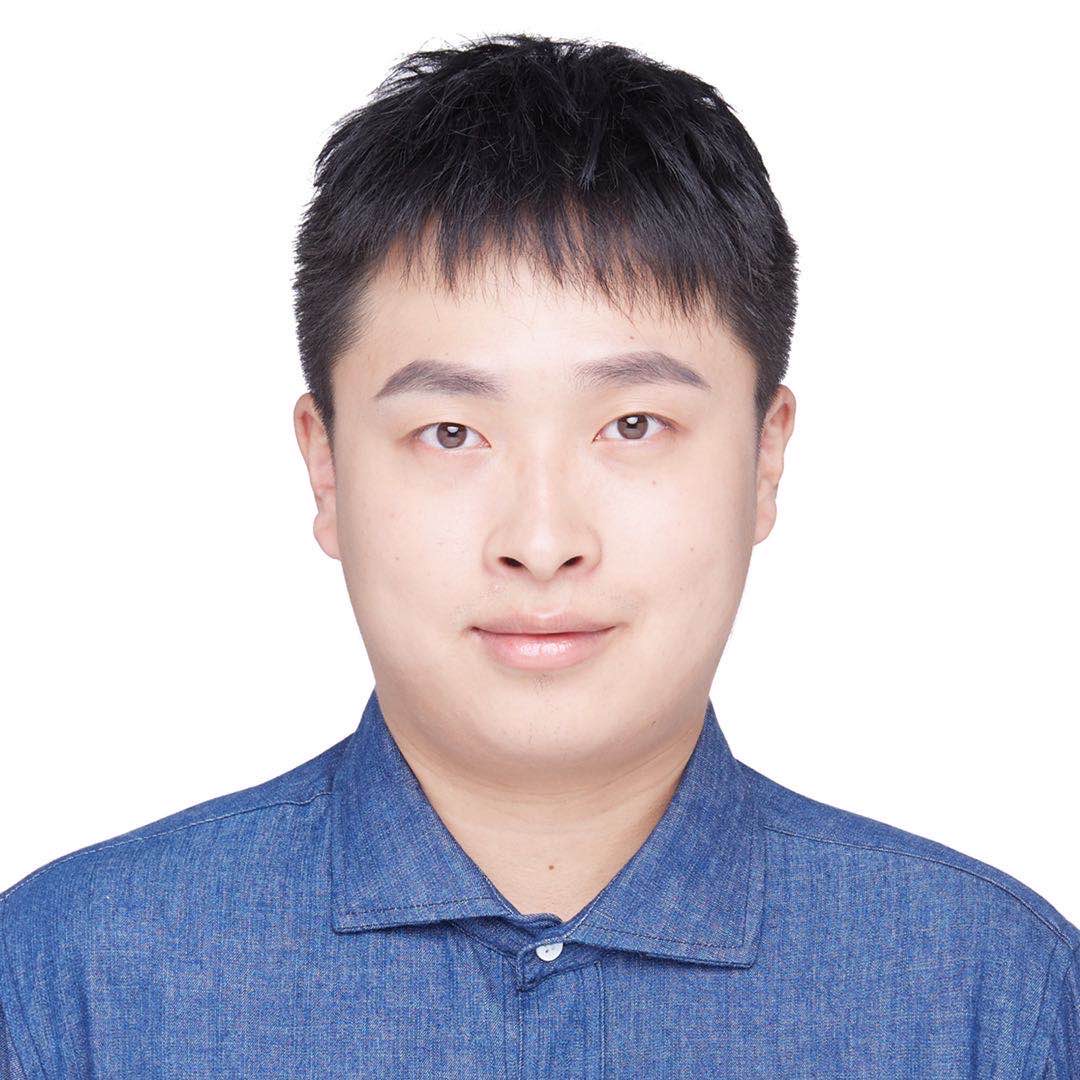}}]
{Weiqi Chen} received his B.Eng. degree in Computer Science and Technology from Xi’an Jiaotong University, China, in 2017 and his M.S. degree in computer science from Zhejiang University, China, 2021. His research interests include urban computing and time series modeling.
\end{IEEEbiography}

\begin{IEEEbiography}[{\includegraphics[width=1in,height=1.25in,clip,keepaspectratio]{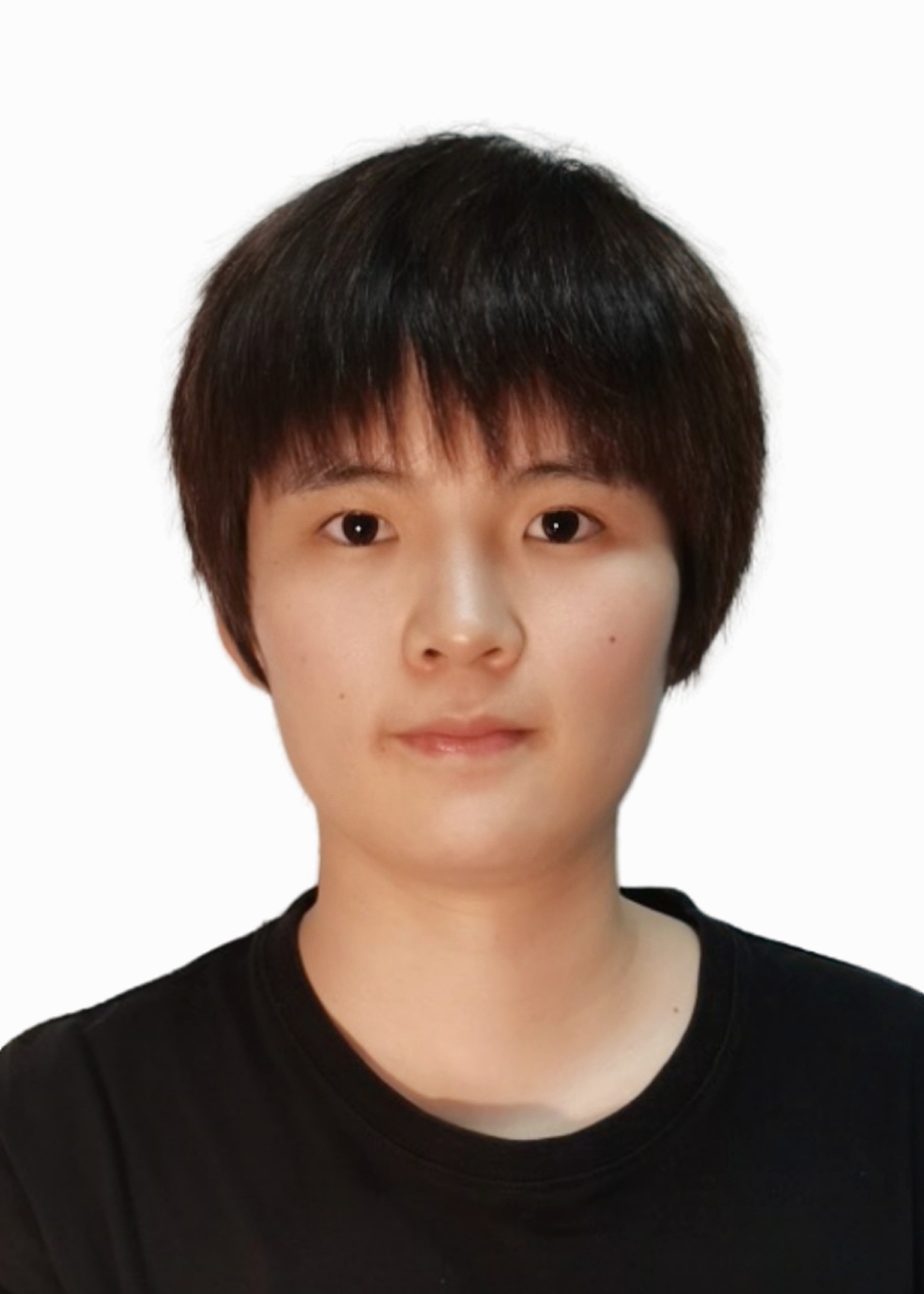}}]
{Binqing Wu} received her B.Eng. degree in computer science from Southwest Jiangtong University, China, in 2020. She is currently a Ph.D. student with the College of Computer Science and Technology, Zhejiang University, China. Her research interests include urban computing and data mining.
\end{IEEEbiography}

\begin{IEEEbiography}[{\includegraphics[width=1in,height=1.25in,clip,keepaspectratio]{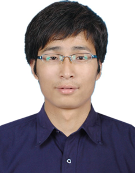}}]
{Youdong Zhang} received his M.A. degree in computer science from Huazhong University of Science and Technology, China, in 2012. He is currently a staff engineer in Alibaba Group. His research interests include distributed storage and database.
\end{IEEEbiography}

\begin{IEEEbiography}[{\includegraphics[width=1in,height=1.25in,clip,keepaspectratio]{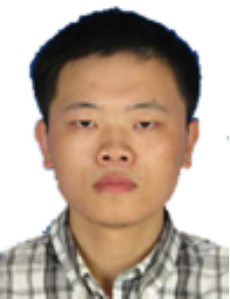}}]
{Bo Wen} received his B.S. degree in computer science from Jishou University, China, in 2012. He is currently a software engineer in Alibaba Group. His research interests include time-series analytics and time-series oriented database.
\end{IEEEbiography}

\vspace{-150 mm}
\begin{IEEEbiography}[{\includegraphics[width=1in,height=1.25in,clip,keepaspectratio]{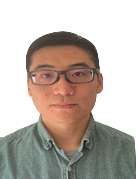}}]
{Chenghu Yang} received his B.S. degree in computer science from Northeast Forestry University, China, in 2009. He is currently a senior staff engineer and leads the NoSQL-Database team in Alibaba Group. His research interests include distributed storage and database
\end{IEEEbiography}
\end{document}